%% file: aaai24.tex
\documentclass[letterpaper]{article} 
\usepackage{aaai24}  
\usepackage{times}  
\usepackage{helvet}  
\usepackage{courier}  
\usepackage[hyphens]{url}  
\usepackage{graphicx} 
\usepackage{natbib}  
\usepackage{caption} 
\usepackage{algorithm}
\usepackage{algorithmic}

\usepackage{amsmath}
\usepackage{booktabs}
\usepackage{multirow}
\usepackage{bigstrut}
\usepackage{soul}
\usepackage[capitalize]{cleveref}
\crefname{section}{Sec.}{Secs.}
\Crefname{section}{Section}{Sections}
\Crefname{table}{Table}{Tables}
\crefname{table}{Tab.}{Tabs.}

\usepackage{newfloat}
\usepackage{listings}
\DeclareCaptionStyle{ruled}{labelfont=normalfont,labelsep=colon,strut=off} 
\floatstyle{ruled}
\newfloat{listing}{tb}{lst}{}
\floatname{listing}{Listing}
\title{Existence Is Chaos: Enhancing 3D Human Motion\\ Prediction with Uncertainty Consideration}
\author{
    Zhihao Wang\textsuperscript{\rm 1,2}, 
    Yulin Zhou\textsuperscript{\rm 1,2}, 
    Ningyu Zhang\textsuperscript{\rm 1}, 
    Xiaosong Yang\textsuperscript{\rm 3}, 
    Jun Xiao\textsuperscript{\rm 1}, 
    Zhao Wang\textsuperscript{\rm 2}\thanks{Corresponding Author.}
}
\affiliations{
    \textsuperscript{\rm 1}Zhejiang University\\
    \textsuperscript{\rm 2}Ningbo Innovation Center, Zhejiang University\\
    \textsuperscript{\rm 3}National Centre for Computer Animation, Bournemouth University\\
    zhao\_wang@zju.edu.cn, zhihao\_wang@zju.edu.cn
}

\usepackage{bibentry}

\begin{document}

\maketitle

\begin{abstract}
Human motion prediction is consisting in forecasting future body poses from historically observed sequences. It is a longstanding challenge due to motion's complex dynamics and uncertainty. Existing methods focus on building up complicated neural networks to model the motion dynamics. 
The predicted results are required to be strictly similar to the training samples with $L_2$ loss in current training pipeline. However, little attention has been paid to the uncertainty property which is crucial to the prediction task. We argue that the recorded motion in training data could be an observation of possible future, rather than a predetermined result. In addition, existing works calculate the predicted error on each future frame equally during training, while recent work indicated that different frames could play different roles. In this work, a novel computationally efficient encoder-decoder model with uncertainty consideration is proposed, which could learn proper characteristics for future frames by a dynamic function. Experimental results on benchmark datasets demonstrate that our uncertainty consideration approach has obvious advantages both in quantity and quality. Moreover, the proposed method could produce motion sequences with much better quality that avoids the intractable shaking artefacts. 
We believe our work could provide a novel perspective to consider the uncertainty quality for the general motion prediction task and encourage the studies in this field. The code will be available in \url{https://github.com/Motionpre/Adaptive-Salient-Loss-SAGGB}.
\end{abstract}

\section{Introduction}
Humans have the ability to predict how an action could be extrapolated in the future. This enables humans to react timely while interacting with external world. In the field of artificial intelligence, how to enable machines to anticipate human behaviour is a paramount challenge. Whether the challenge is handled well affects many real-life applications such as autonomous driving~\cite{paden2016survey, djuric2020uncertainty} and human-robotics interaction~\cite{koppula2013anticipating}.
The task of human motion prediction can be described as giving a series of human pose sequences in the past and predicting future human pose sequences. Anticipating the future movement of the 3D human skeleton is very challenging due to the complex spatial-temporal modality and the great uncertainty of the future.

\begin{figure}[t]
  \includegraphics[width=\columnwidth]{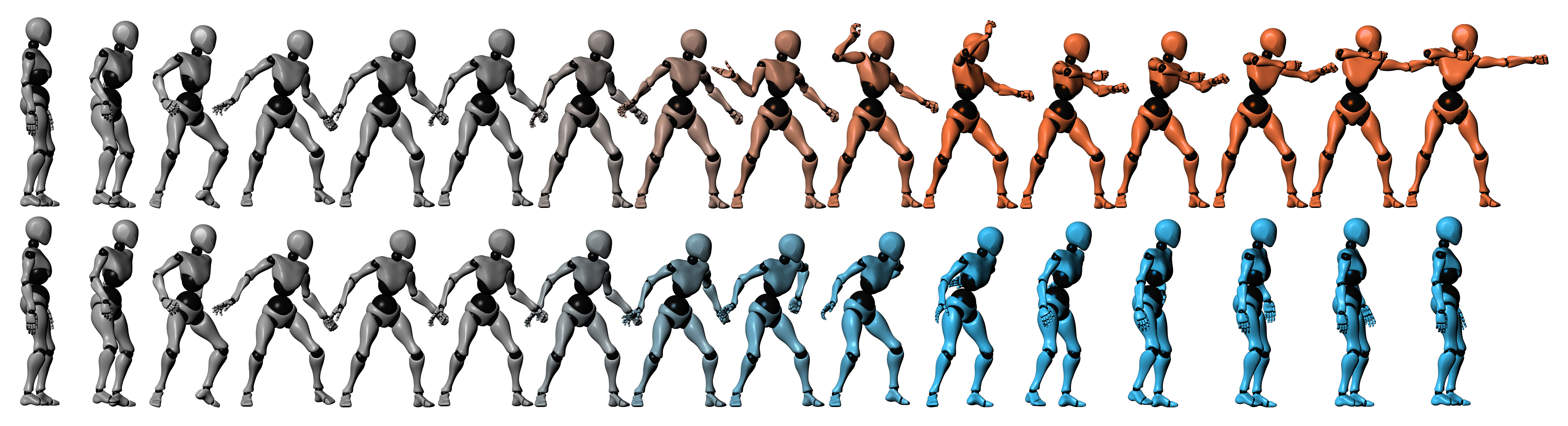}
  \caption{The uncertainty characteristic of human motion. For a certain motion clip, the recorded future motion should be an observation of possible future, rather than a predetermined result. The future motion could vary even with similar historical motion. }
  \label{fig:teaser}
\end{figure}

\begin{figure}[htbp]
\centering
\includegraphics[width=0.9\columnwidth]{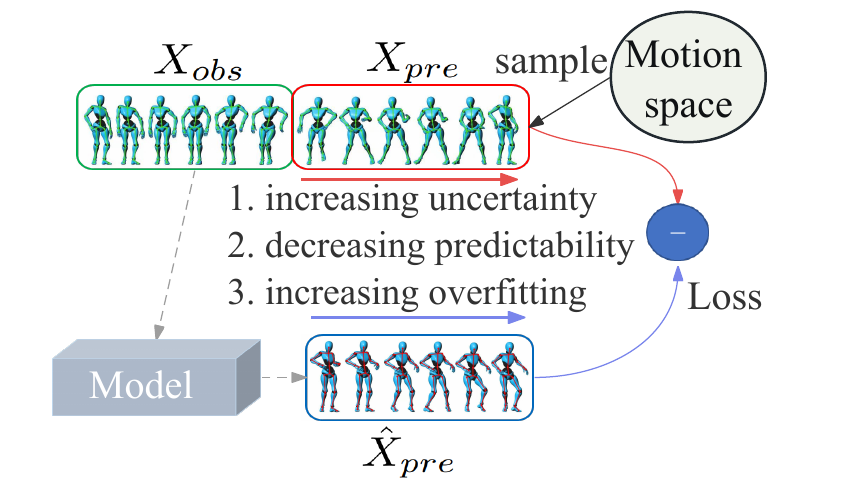} 
\caption{Training pipeline of existing prediction approaches. A given piece of training data sample would be divided into past pose sequences $X_{obs }$ and future pose sequences  $X_{pre }$. The predicted result $\hat{X}_{pre }$ is required to be strictly similar to the training data $X_{pre }$, where the uncertainty of future motion has been ignored. }
\label{fig:train_pipeline}
\end{figure}

Current popular training pipeline of existing approaches is shown in \cref{fig:train_pipeline}. A given piece of training data sample would be divided into past pose sequences and future pose sequences $\{X_{obs },X_{pre }\}$. The module would generate predicted future sequence $\hat{X}_{pre }$ while the parameters would be updated with the gradient of ${loss}(X_{pre ,}\hat{X}_{pre})$. 
The prediction results are required to be strictly similar to the training examples. However, the uncertainty property of motion is overlooked. It would meet difficulties while dealing with the scenario that different motions may begin with the similar human motion poses. Thus, it is critical to take the uncertainty into consideration in human motion prediction. 

The uncertainty of motion prediction mainly refers to its challenging variation, e.g. a difficult pose or a distant frame, especially for non-periodic behaviors.  
Most existing approaches treat each future frame equally.  However, the uncertainty consideration of human motion is actually not equal in each future frame. In our real-life experience, the short-term future of a given action is much easier to be predicted 
, but the possible ground truth for movements in the longer period could be diverse. Such diverse could be varied due to the type of actions but generally become larger over time. In other word, the recorded future sequences in a motion clip should be an observation of historical sequence's possible future, rather than a predetermined result. For example, humans could predict a walking action will lift their feet off ground in the next millisecond, however, humans can walk to the left side or walk to the right side a few minutes later.  The inherent relationships between the previous frames and longer future frames will generally go weaker over time. 

Motivated by such insight, we aim to explicitly model the uncertainty of human motion for achieving high precision and robustness prediction performance. We challenge the widely used average approaches that do not differentiate the weight of each frame. The proposed model is encouraged to focus more on learning the credible frames accurately. 
Two significant pieces of evidence from existing empirical results support our idea: (1) Long-term error accumulation has been recognized as one of the biggest issues which bring performance degradation in motion prediction problems. As most models will predict the next frame conditioned on the previously predicted sequence, a small error in the initial frame will be amplified greatly due to the butterfly effects. (2) The latest work~\cite{ma2022progressively} proves that a different initial pose could bring sharp performance gains than an individual method. 
These validate that a more accurate prediction of the early frames will matter a lot in the final results. 

Moreover, the long-term prediction from observation would usually have low confident, and this could vary due to the motion type. In light of these findings, a self-attention graph generate block (SAGGB) is designed to leverage the certainty connection information from the diversity of actions and the complexity of behaviors . Additionally, an Adaptive-Salient Loss is presented to make effective use of the uncertainty property to produce realism long-term prediction results. We consider properly assigning the weight to each frame, and present an active function to dynamically learn the weights for frames. 

To evaluate our idea, 
extensive experiments are carried out on H3.6M, 3DPW and CMU Mocap datasets to study the impact of learning early frames for the final performance. The results demonstrate that our method achieve competitive performance in both short-term and long-term motion prediction tasks. Besides, our prediction results are more smooth and natural, achieve high quality without the intractable shaking effects. 

In summary, the contributions of this paper are the followings:
\begin{enumerate}
    \item  The role of the uncertainty property in human motion prediction tasks has been studied, where its importance, and subsequently elaborated on its mechanism and principles are revealed. We hope our work will encourage more studies to rethink the value of uncertainty factors in the motion prediction problems.
    \item  A novel motion prediction work involved uncertainty consideration is proposed. A dynamic function is designed to learn the assigned weights for future frames. Extensive experimental results on the benchmark datasets validate our method outperform competitors in most short-term prediction jobs and achieve competitive performance in long-term prediction jobs. The proposed method could bring more favourable gains than existing methods in both quantity and quality. 
    \item We delve deeper into the uncertainty assumption and carried out extensive experiments to pursue the problem of how could the different assigned weights could affect the training and learning of the motion prediction models. Fruitful insights are given out by our ablation studies.
\end{enumerate}

\begin{figure*}[!htbp]
\centering
\includegraphics[width=0.95\textwidth]{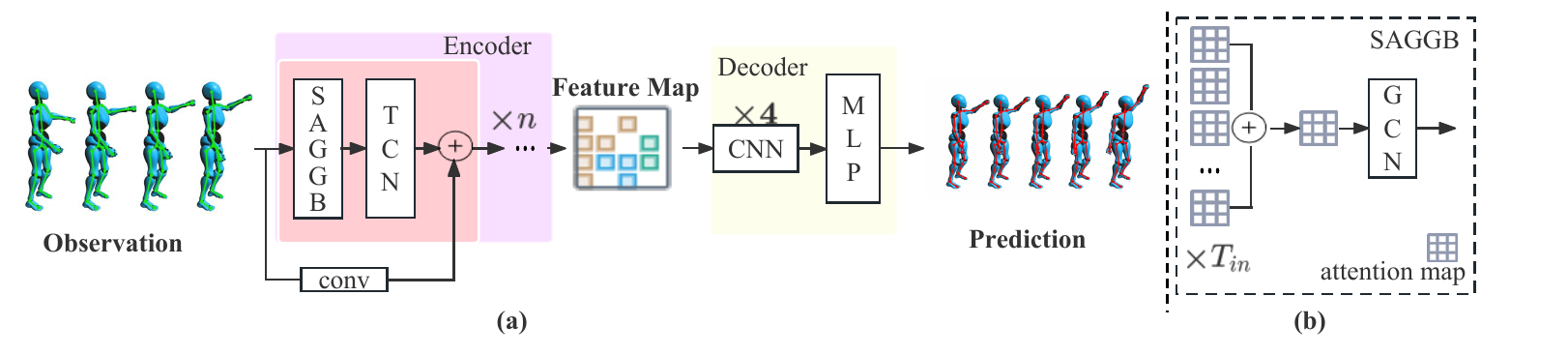} 
\caption{Overview of proposed model for human motion prediction with SAGGB. In the encoder, SAGGB leverage self attention mechanism to generate sample related graph to extract spatial information. In the decoder, we use lightweight CNNs and MLP to predict.  }
\label{fig:architecture}
\end{figure*}

\section{Related Work}
\subsection{Human Motion Prediction}
Recently, deep learning networks became the mainstream of motion prediction. Since motion prediction was often seen as a sequence-to-sequence task, RNN-based methods were naturally used for this task. For example, LSTM-3LR~\cite{jain2016structural} and ERD~\cite{fragkiadaki2015recurrent} introduced LSTM into the task. ERD added encoder and decoder before and after LSTM to achieve better results. LSTM-3LR replaced bone information with speed information. However, RNNs suffer from error accumulation and were unable to effectively model spatial information.
Then feed-forward neural networks were applied to the field. They tried to use convolution kernels to extract information in space and time. TE~\cite{butepage2017deep} and QuaterNet~\cite{pavllo2018quaternet} used CNNs to capture temporal information. convSeq2Seq~\cite{li2018convolutional} and CHA~\cite{li2019efficient} designed hierarchical CNNs to model spatial and temporal information simultaneously. The TrajectoryCNN~\cite{liu2020trajectorycnn} proposed the features of trajectory space that were more easily handled by CNN. The siMLPe~\cite{guo2023back} proposed to use simple multi-layer MLPs to extract temporal and spatial information and achieves excellent results. Some probabilistic prediction works~\cite{yuan2020dlow, mao2022weakly} predicted multiple possible motion sequences. However, due to the different evaluation metrics, our method does not compare with these works.

\subsection{Graph Convolutional Networks (GCNs)}
Recently, GCNs achieved the state-of-art results in motion prediction task. LTD~\cite{mao2019learning} saw pose graph as a fully-connected graph and modeled temporal information by DCT representations which were followed by lots of later work. JDM~\cite{su2021motion} and MPT~\cite{liu2021motion} proposed GCNs that took velocity information as input. In the graph representation, DMGNN~\cite{li2020dynamic} and MSR-GCN~\cite{dang2021msr} used muti-scale graph and extracted information from single scales and cross scales. STSGCN~\cite{sofianos2021space} and GAGCN~\cite{zhong2022spatio} built space-time separable GCN to extract information in graph. SPGSN~\cite{li2022skeleton} did scattering decomposition of the graph and GCNs were performed on all graphs. In the learning strategy, PGBIG~\cite{ma2022progressively} performed muti-stage prediction whose output corresponded to a smooth sequence in every stage. UA-HMP~\cite{ding2021uncertainty} integrated probabilistic prediction into deterministic methods. AuxFormer~\cite{xu2023auxiliary} introduced a model learning framework with auxiliary task which was recovering corrupted coordinates depending on the rest coordinates. DMAG~\cite{gao2023decompose} used frequency decomposition and feature aggregation respectively to encode the information. Meanwhile, Transformer was also attempted for human motion prediction~\cite{mao2020history,cai2020learning,aksan2021spatio}. Self-attention mechanism of Transformer had strong modeling ability on sequence data and was able to model the dependencies of joints. In our model, we synthesize the modeling ability of self-attention mechanism and the effectiveness of graph neural networks.


\section{Methodology}
\subsection{Overview}
Firstly, some notations and variables in this paper would be enumerate. We denote $X_{obs }=[x_{1}, x_{2},\cdot, x_{T_{in} }] $ as the past pose sequences and $X_{pre }=[x_{T_{in}+1}, x_{T_{in}+2},\cdot, x_{T_{in}+T_{out} }] $ as the future pose sequences, where $ x_{i}\in  R^{N\times d} $ with N joints and d-dimension space (d is 3) is  represented as human pose at time i. The goal of motion prediction is to learn a mapping function that maps $X_{obs }$ to $X_{pre }$. 

The overview of proposed model is showed in \cref{fig:architecture}. An encoder-decoder structure is adopted to conduct end-to-end prediction.
First, an encoder with proposed Self-attention Graph Blocks (SAGGBs) and Temporal Convolutional Network modules (TCNs) would encode the input into high-dimensional feature spaces. Then a lightweight decoder makes predictions for future pose sequence estimation.

The encoder contains a series of residual blocks. Each block consists of SAGGB and TCN which respectively extract the spatial and temporal information. After feature extraction through all blocks, the encoder would output a feature map of $R^{T_{in}\times V \times C_f}$, where $C_f=128$ is used in this work. 
The decoder is a lightweight design that includes 4 CNNs and a MLP. The CNN module uses time dimension as feature channel to extract information. The first CNN projects the input from $R^{T_{in}\times V \times C_f}$ to $R^{T_{out}\times V \times C_f}$. The MLP projects the $C_f$ to 3.

\subsection{Self-attention Graph Generate Block} For GCNs, either natural connection or full connection applies the same graph weights among all samples, where the the diversity of actions and the complexity of behaviors are ignored. For instance, for action walking, the dependence between  feet should be more of a concern but for action eating the dependence is not so important. Such dependency reflects the variation of uncertainty with motion diversity.  Hence, We propose a Self-attention Graph Generate Block (SAGGB) module to generate data-driven graph to model such information. Let's recall $x_{i} \in R^{n\times d} $ as the human pose at time $i$. We apply self-attention mechanism to generate graph $A_{i}$ for every pose

\begin{equation}
    A_{i} = softmax(\frac{\sigma (x_{i})\phi (x_{i}) ^\mathrm{T}  }{\sqrt{d_{k}}}) 
\end{equation}
where $\sigma ( .) $ and $\phi(.)$ are mapping function to generate query and key for every joint; $d_{k}$ is the dimension of query and key. For each pose, SAGGBs generate attention map about the joints as the graph to this pose. As can be noticed, SAGGBs generate different graphs with different weights for different poses.

However, the same pose in different pose sequences could have different means. Consequently, it's more sensible to generate special graph for every pose sequence rather than pose. We denote the graph $A_{sample}$ for a given sample as below
\begin{equation}
    A_{sample} = \sum_{i=1}^{T_{in} } A_{i}  
\end{equation}

By constructing a new graph for each sequence, the model is able to extract implicit higher-dimensional information from different input poses.Based on the special graph for each sample, the output $X^{l+1}$ of SAGGB block with given input $ X^{l} $ can be defined as:
\begin{equation}
    X^{l+1} = \sigma (A_{sample}X^{l}W^{l})
\end{equation}

\input{Table_h36_short}

\subsection{Loss Function}
To train the model for human motion prediction, $L_2$ loss is usually used in existing approaches:
\begin{equation}
    \ell = \frac{1}{T_{out}N } \sum_{t=1}^{T_{out} }\sum_{n=1}^{N} \left \| \hat{x }_{t}^{n}-x_{t}^{n}   \right \|_{2}    
\end{equation}
As mentioned before, such averaging form treat all timestamps equally, where problem arises in two aspects. First, the prediction of the later frames could be difficult due to the uncertainty of the action semantics.  
However, the loss with averaging form gives more weight to the later frames with greater uncertainty and less structured knowledge, which makes the model fit more noise and uncertainty. Second, human motion is a continuous process, thus the distribution of one frame is heavily dependent on the state of the previous frame. This dependency is passed frame by frame. Therefore, the prediction of the first frame in the prediction sequence is crucial. The first frame in prediction determines the initial position of the model from the deterministic space to the uncertain space. Meanwhile, the dependency relationship shows that the loss of the first frame prediction largely determines the expression of the entire model.

Based on the above two issues, we propose an Adaptive-Salient Loss in this work.
\paragraph{Adaptive Loss} Although motion uncertainty varies in different actions, it mainly increases over time. To overcome this imbalance problem, inspired by multi-task model~\cite{kendall2018multi}, we propose an Adaptive Loss for human motion prediction. 
First, for the output of the model $\hat{X}_{pre }$, we define a probabilistic model as:
\begin{equation}
p\left ( x_{t} |\hat{x}_{t}  \right ) = \mathcal{N}\left ( \hat{x}_{t}, \sigma  \right ) 
\end{equation}
\begin{equation}
p\left ( \left [ x_{1}, x_{2}, \dots ,x_{T_{out}}  \right ]|[\hat{x_{1}}, \hat{x_{2}}, \dots ,\hat{x}_{T_{out}}]     \right ) = \prod_{t=1}^{T_{out}}p(x_{t}|\hat{x_t})
\end{equation}
where $\sigma$ denotes the uncertainty. 
To optimize it, we maximise the log likelihood.
\begin{equation}
    \log{p\left ( x_{t} |\hat{x_{t} }  \right )} \propto -\frac{1}{2\sigma ^{2} }\left \| x_{t} - \hat{x_{t} }   \right \|_{2} - \log{\sigma }  
\end{equation}
Consequently, we regard the prediction for each frame as a part of task and combine the tasks to get better predictions on the whole.
\begin{equation}
    \log{p\left ( X_{pre }|\hat{X}_{pre }   \right )  }  \propto - \sum_{t=1}^{T_{out}}(\frac{1}{2\sigma _{t}^{2} } \left \| x_{t} - \hat{x_{t} }   \right \|_{2}+\log{\sigma_{t} })   
\end{equation}
Accordingly, we can define the proposed Adaptive Loss as \cref{eq9}.
Adaptive Loss reflects the increasing of uncertainty over time. At prediction sequences, $\sigma$ is the increasing uncertainty over the time. During training, $\sigma$ adjusts the weights of different frames hence model gives variant attention to frames.

\begin{equation}
    L_{adaptive} = \sum_{t=1}^{T_{out}}\frac{1}{2\sigma _{t}^{2} }\left \| x_{t}-\hat{x_{t} }   \right \|_{2} +\sum_{t=1}^{T_{out}}\log{\sigma _{t} }
    \label{eq9}
\end{equation}

\paragraph{Salient Loss} In Adaptive Loss, we build the probabilistic model based on the assumption that the prediction for each frame is independent. However, human motion is a continuous process and the distribution of one frame is heavily dependent on the state of the previous frame. We formulate the perception as follows:

\begin{equation}
    x_{t+1} = f(x_t,\sigma_{t+1},\upsilon _{t})
    \label{eq10}
\end{equation}
where $\upsilon_{t}$ is temporal and motion information before timestamp t. $\sigma_{t+1}$ represent uncertainty spaces in $t+1$.

\cref{eq10} is intuitive that expresses the continuity and uncertainty of human motion. The continuity is also essential while making predictions. Therefore, as the initial state of the model output prediction sequence, the last input frame is crucial to the entire prediction process, which can be expressed as:

\begin{equation}
    \hat{X}_{pre} = F(x_{T_{in}},\sigma_{T_{in}+1:{T_{in}+T_{out}}}, \upsilon _{T_{in}: {T_{in}+T_{out}-1}})
    \label{f1}
\end{equation}

The last input frame is the initial state of the prediction output in the deterministic space. \cref{f1} expresses that the prediction process needs to explicitly utilize the last input frame. Many previous methods have taken advantage of this. In LTD~\cite{mao2019learning} and SPGSN~\cite{li2022skeleton}, the last input frame is repeated $T_{out}$ times after input as the input of the model, and the offset relative to the last frame is output. PGBIG \cite{ma2022progressively} repeats the last frame $T_{out}$ times as an initial estimate for the predicted sequence.

Meanwhile, the first predicted frame is the initial state of the prediction output in the uncertainty space. Hence, we can continue the progressive deduction of prediction and reformulated \cref{f1} as:

\begin{equation}
    \hat{X}_{pre} = F(\hat{x}_{T_{in}+1},\sigma_{T_{in}+2:{T_{in}+T_{out}}}, \upsilon _{T_{in}+1: {T_{in}+T_{out}-1}})
    \label{f2}
\end{equation}

Previous methods explicitly utilize the initial state of the output in the deterministic observed data as shown in \cref{f1}. The initial state of the model in the prediction with uncertainty is ignored.  In order to tackle this issue, we propose a Salient Loss as follows:

\begin{equation}
    L_{salient} = \omega T_{out} \left\| x_1 - \hat{x}_1 \right\|_{2} +  \sum_{t=1}^{T_{out}}\left \| x_{t}-\hat{x}_{t}   \right \|_{2}
\end{equation}
where $\omega$ is salient factor. 

\input{Table_h36_long}

\paragraph{Adaptive-Salient Loss} Adaptive Loss indicates that early stage prediction usually has higher confident. During training, we treat $\sigma$ as a set of trainable parameters and $\sigma$ is learned together with the model parameters. The $\log{\sigma}$ is the regularization term. 

Salient Loss could emphasize first frame's importance as the initial state of the prediction sequence with uncertainty. During training, we set $\omega$ as a fixed value to emphasize the importance of the initial pose.

Our final loss function is a weighted combination of Adaptive Loss and Salient Loss:
\begin{equation}
    L = \lambda L_{Adaptive} + (1-\lambda)L_{Salient}
\end{equation}
where $\lambda$ is weight factor.

\section{Experiments}


\subsection{Datasets}
\label{sec:results:dataset}
\textbf{Human3.6M dataset (H3.6M)} is a widely used motion prediction dataset consists of 15 actions performed by 7 actors(S1,S5,S6,S7,S8,S9 and S11). The human body is represented as 32 joints. Follow~\cite{li2022skeleton,ma2022progressively}, global information including global rotation and movement is removed. Meanwhile, all samples are down-sampled to 25 frames per second and we test our model on S5.

\noindent\textbf{CMU Motion Capture  dataset (CMU Mocap)} is another dataset widely used for human motion prediction. Like previous methods, we use 8 actions and choose 25 joints for each pose. Other processing is similar to H3.6M.

\noindent\textbf{3D Pose in the Wild dataset (3DPW)} includes both indoor and outdoor actions captured at 30Hz. Each pose has 26 joints and we use 23 of them. We conduct experiments according to the official segmented training set, validation set and test set.

\subsection{Evaluation Criterion and Baselines} 
\label{sec:results:baseline}
Follow ~\cite{ma2022progressively}, Mean Per Joint Position Error (MPJPE) is employed as criterion which calculates the average $L_2$ distance at each timestamp. 


The proposed method is compared with 
DMGNN~\cite{li2020dynamic}, LTD~\cite{mao2019learning}, SPGSN~\cite{li2022skeleton} and PGBIG~\cite{ma2022progressively}. All methods are tested under the same conditions. Meanwhile, the proposed method is also compared with STSGCN~\cite{sofianos2021space} and GAGCN~\cite{zhong2022spatio} in H3.6M under their evaluation criteria. Inspired by ~\cite{du2023avatars}, we further employ Jitter metric to evaluate the quality of predicated motion sequences. 
\begin{equation}
    \text{Jitter}=\frac{\mu^3}{T-3} \sum_{t=0}^{T-3} (\Delta x_{t+3} - 3*\Delta x_{t+2} + 3*\Delta x_{t+1} - \Delta x_{t}  )
\end{equation}
where $\Delta x_{t}$ is the Euclidean distance between prediction pose and future pose at frame $t$, $T$ is the total frames and $\mu$ is the frame per second. 

\subsection{Comparison with the State-of-the-art Methods}
\label{sec:results:sota}

To show the performance of our model, we list the quantitative results for both short-term prediction(400ms) and long-term prediction(1000ms) on H3.6M, CMU Mocap and 3DPW.
\input{Table_h36_stsgcn.tex}
\begin{figure*}[htbp]
\centering
\includegraphics[width=0.95\textwidth]{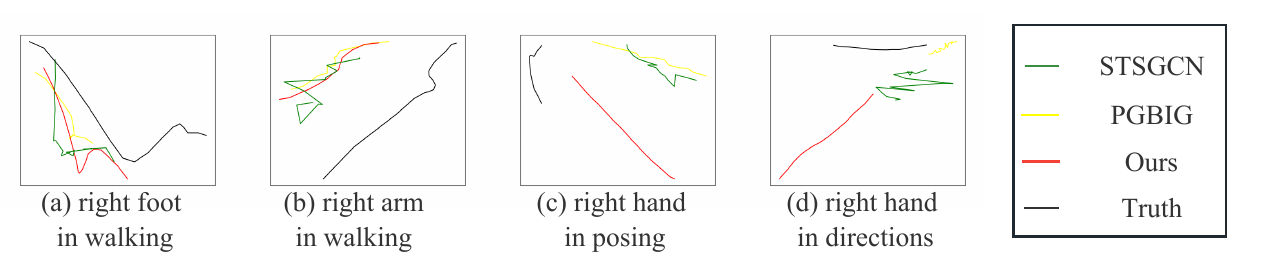} 
\caption{Comparison of Qualitative Evaluation Results. Joint trajectory is recorded from 400ms to 1000ms. The shaking problems have occurred in comparison method's prediction results, which could be caused by over-fitting.}
\label{fig3}
\end{figure*}

\paragraph{H3.6M} \cref{h36_short} shows the short-term comparisons results. Our model outperforms state-of-art methods in most cases of all actions, which shows good stability compared to other methods by taking into account uncertainty and better encoding spatial information. 

The long-term comparisons is shown in \cref{h36_long}. Our model preferentially guarantees the optimization of the early frame. It also performs competitively in the long-term prediction due to less fitting noise and uncertainty. Meanwhile, although current state-of-art methods methods could provide impressive long-term prediction MPJPE score, the motion quality of predicted results remain in doubt. 
This part would be detail discussed in the following paragraph \ref{sec:exp:QA} and an example video is illustrated in the supplementary material. 

\input{Table_cmu}
\input{Table_3dpw}

\paragraph{CMU Mocap \& 3DPW} \cref{cmu} and \cref{3dpw} show the comparison on CMU Mocap and 3DPW with both short-term and long-term prediction. Only average error is listed due to space limitations. We also achieve best in most cases, which demonstrates the effectiveness of our method.


\paragraph{Computational Complexity Analysis} The computational complexity and time consuming results are  shown in \cref{complexity}, where the model checkpoint provided by original public repository is used. Our model is smaller than previous methods as we conduct a lightweight architecture. Although the calculation cost of our model is increased due to the self-attention, the actual inference time of our model is still in advantage.

\input{Table_complexity.tex}

\begin{figure}[b]
\centering
\includegraphics[width=0.95\columnwidth]{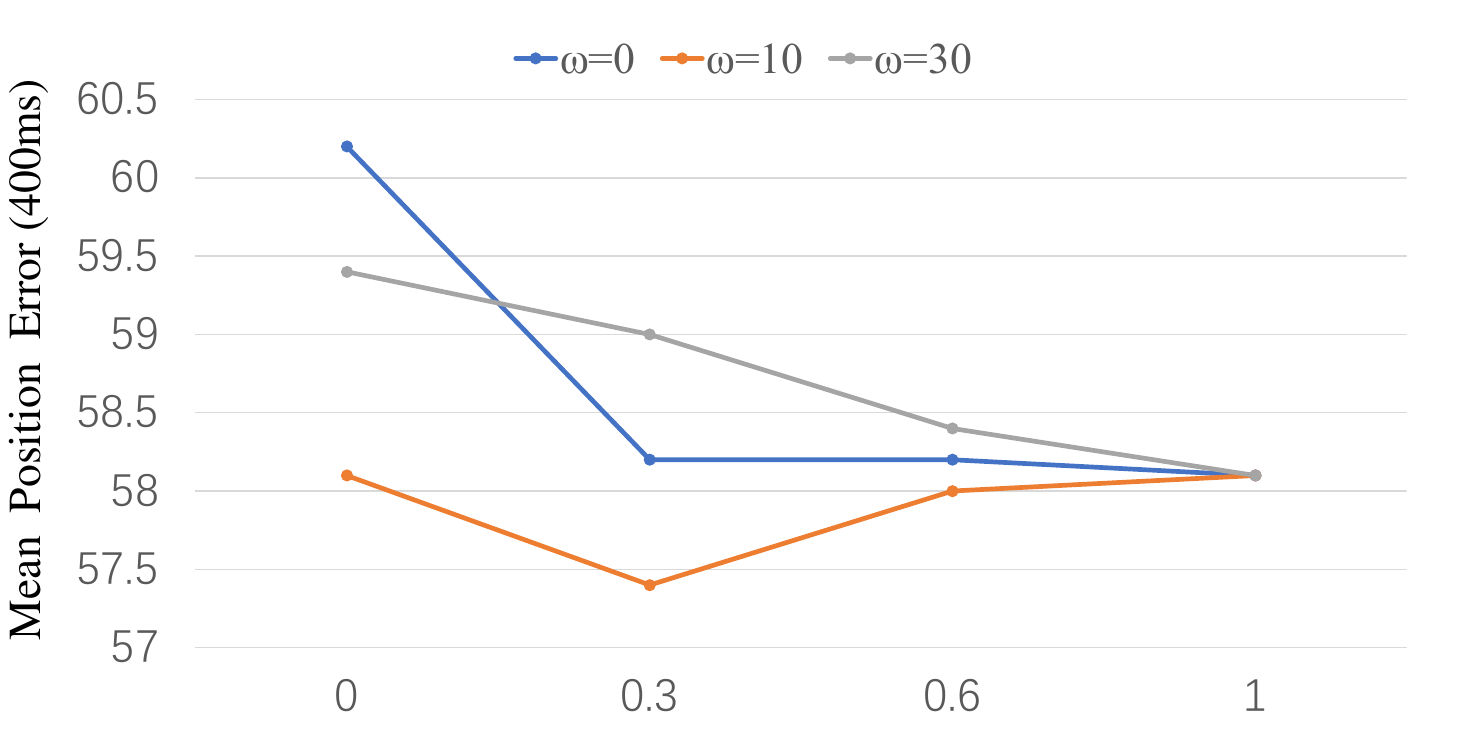} 
\caption{Ablation on value of $\lambda$ and $\omega$ on H3.6M}
\label{fig4}
\end{figure}

\subsection{Quality Analysis of Predicted Motion}
\label{sec:exp:QA}
Statistical analysis on prediction quality is shown in \cref{jitter} using Jitter metric \cite{du2023avatars}. To be specific, the sequences predicted by our method performed better in realism and stable with the increase of prediction frames. Visualization of Joint Trajectory is shown in \cref{fig3}. The trajectory of a single joint point projected on the x-y plane of a prediction sequence is shown. More turns appear at the end of the PGIBIG's and STSGCN's predictions to mimic the real joint motion. Comparing to baselines, ours predicted results are more reasonable and smooth. Further details in prediction result is shown in the supplementary video. 

\input{Tabel_jitter}

\input{Table_ablation_model.tex}

\subsection{Ablation Study}  The ablation study results for hyper parameters is shown in \cref{fig4}. These results confirm the validity of our proposed loss and also support our hypothesis. The salient factor $\omega$ and weight factor $\lambda$ is determined as $10$ and $0.3$, respectively.
In addition, we also investigate the validity of our model structure. We have conducted the ablation study of proposed SAGGB and investigate the block number for encoder, where the result is shown on \cref{ablation_model}. 

\subsection{Discussions}
Our method achieves high quality and realism motion prediction. However, there are still two limitations:
1. The popular $L_2$ Loss, e.g. MPJPE may be not sufficient enough to fully express the quality of the prediction and be an indicator for a prediction model. Especially long-term prediction results with similar MPJPE could perform quite different on motion stable and realism. Further evaluation indicators could be conducted to improve model training and prediction quality evaluation. 
2. Current benchmark datasets are originally created for either pose estimation or action recognition purpose. Hence, some key characteristics of motion prediction are not considered well, e.g. various possible action modes from various subjects.  In the future, how to construct a dataset that is more suitable for human action prediction is worthy of future research.

\section{Conclusion}
To sum up, we have proposed a novel motion prediction framework with uncertainty consideration, which challenges the assumption that training data should be fully trusted equally. 
Extensive studies have been carried out to evaluate our insights on the heavily benchmark Human3.6M, 3DPW and CMU datasets. Our method could achieve favourable gains. More importantly, our method could tackle over-fitting problem that avoids weird artifacts and generates more realistic motion sequences. Extensive ablation studies are carried out to present fruitful insights into the field. We believe this new perspective of uncertainty will inspire other researchers and facilitate the prediction related work both in academic and industries in the future.

\newpage
\section{Acknowledgments}
This research has been supported by National Key Research and Development Project of China (Grant No. 2021ZD0110702), Natural Key Research and Development Project of Zhejiang Province (Grant No. 2023C01043), Ningbo Natural Science Foundation (Grant No. 2023Z236)
, Information Technology Center and State Key Lab of CAD\&CG, Zhejiang University.
\bibliography{aaai24}

\end{document}

%% file: Table_h36_short.tex
\begin{table*}[t]
\small
\addtolength{\tabcolsep}{-2pt}
  \centering
    \begin{tabular}{c|cccc|cccc|cccc|cccc}
    \hline
    action & \multicolumn{4}{c|}{walking} & \multicolumn{4}{c|}{eating} & \multicolumn{4}{c|}{ smoking} & \multicolumn{4}{c}{discussion} \bigstrut[t]\\
    \hline
    millisecond & 80   & 160  & 320  & 400  & 80   & 160  & 320  & 400  & 80   & 160  & 320  & 400  & 80   & 160  & 320  & 400 \bigstrut[t]\\
    \hline
    DMGNN & 17.3  & 30.7  & 54.6  & 65.2  & 11.0  & 21.4  & 36.2  & 43.9  & 9.0  & 17.6  & 32.1  & 40.3  & 17.3  & 34.8  & 61.0  & 69.8  \bigstrut[t]\\
    LTD  & 12.3  & 23.0  & 39.8  & 46.1  & 8.4  & 16.9  & 33.2  & 40.7  & 7.9  & 16.2  & 31.9  & 38.9  & 12.5  & 27.4  & 58.5  & 71.7  \bigstrut[t]\\
    SPGSN & 10.1  & 19.4  & 34.8  & 41.5  & 7.1  & 14.9  & 30.5  & 37.9  & 6.7  & 13.8  & 28.0  & 34.6  & 10.4  & 23.8  & 53.6  & 67.1  \bigstrut[t]\\
    PGBIG & 10.2  & 19.8  & 34.5  & 40.3 & 7.0  & 15.1  & 30.6  & 38.1  & 6.6  & 14.1 & 28.2  & 34.7  & 10.0  & 23.8  & 53.6  & 66.7  \bigstrut[t]\\
    \hline
    Ours & \textbf{8.8 } & \textbf{17.7 } & \textbf{33.1 } & \textbf{39.6}  & \textbf{6.2 } & \textbf{14.1 } & \textbf{29.8 } & \textbf{37.3 } & \textbf{5.7 } & \textbf{12.8 } & \textbf{26.9 } & \textbf{33.7 } & \textbf{8.7 } & \textbf{21.8 } & \textbf{51.4 } & \textbf{64.9 } \bigstrut[t]\\
    \hline
    action & \multicolumn{4}{c|}{directions} & \multicolumn{4}{c|}{greeting} & \multicolumn{4}{c|}{phoning} & \multicolumn{4}{c}{posing} \bigstrut[t]\\
    \hline
    millisecond & 80   & 160  & 320  & 400  & 80   & 160  & 320  & 400  & 80   & 160  & 320  & 400  & 80   & 160  & 320  & 400 \bigstrut[t]\\
    \hline
    DMGNN & 13.1  & 24.6  & 64.7  & 81.9  & 23.3  & 50.3  & 107.3  & 132.1  & 12.5  & 25.8  & 48.1  & 58.3  & 15.3  & 29.3  & 71.5  & 96.7  \bigstrut[t]\\
    LTD  & 9.0  & 19.9  & 43.4  & 53.7  & 18.7  & 38.7  & 77.7  & 93.4  & 10.2  & 21.0  & 42.5  & 52.3  & 13.7  & 29.9  & 66.6  & 84.1  \bigstrut[t]\\
    SPGSN & 7.4  & 17.2  & 39.8  & 50.3  & 14.6  & 32.6  & 70.6  & 86.4  & 8.7  & 18.3  & 38.7  & 48.5  & 10.7  & 25.3  & 59.9  & 76.5  \bigstrut[t]\\
    PGBIG & 7.2  & 17.6  & 40.9  & 51.5  & 15.2  & 34.1  & 71.6 & 87.1  & 8.3  & 18.3  & 38.7 & 48.4 & 10.7  & 25.7  & 60.0 & 76.6 \bigstrut[t]\\
    \hline
    Ours & \textbf{6.2 } & \textbf{16.0 } & \textbf{39.0 } & \textbf{50.0 } & \textbf{12.5 } & \textbf{30.4 } & \textbf{68.6}  & \textbf{85.4}  & \textbf{7.4 } & \textbf{17.1 } & \textbf{37.8}  & \textbf{47.9}  & \textbf{9.1 } & \textbf{23.3 } & \textbf{57.4}  & \textbf{74.6}  \bigstrut[t]\\
    \hline
    action & \multicolumn{4}{c|}{purchases} & \multicolumn{4}{c|}{sitting} & \multicolumn{4}{c|}{sittingdown} & \multicolumn{4}{c}{takingphoto} \bigstrut[t]\\
    \hline
    millisecond & 80   & 160  & 320  & 400  & 80   & 160  & 320  & 400  & 80   & 160  & 320  & 400  & 80   & 160  & 320  & 400 \bigstrut\\
    \hline
    DMGNN & 21.4  & 38.7  & 75.7  & 92.7  & 11.9  & 25.1  & 44.6  & 50.2  & 15.0  & 32.9  & 77.1  & 93.0  & 13.6  & 29.0  & 46.0  & 58.8  \bigstrut[t]\\
    LTD  & 15.6  & 32.8  & 65.7  & 79.3  & 10.6  & 21.9  & 46.3  & 57.9  & 16.1  & 31.1  & 61.5  & 75.5  & 9.9  & 20.9  & 45.0  & 56.6  \bigstrut[t]\\
    SPGSN & 12.8  & 28.6  & 61.0  & 74.4  & 9.3  & 19.4  & \textbf{42.3}  & \textbf{53.6}  & 14.2  & 27.7  & 56.8  & 70.7  & 8.8  & 18.9  & 41.5  & \textbf{52.7}  \bigstrut[t]\\
    PGBIG & 12.5  & 28.7  & 60.1  & \textbf{73.3 } & 8.8  & 19.2  & 42.4  & 53.8  & 13.9  & 27.9  & 57.4  & 71.5  & 8.4  & 18.9  & 42.0  & 53.3  \bigstrut[t]\\
    \hline
    Ours & \textbf{10.9 } & \textbf{26.8 } & \textbf{59.8 } & 73.9  & \textbf{8.1 } & \textbf{18.4 } & \textbf{42.3 } & 54.1 & \textbf{12.8 } & \textbf{26.3 } & \textbf{55.9 } & \textbf{70.3 } & \textbf{7.8 } & \textbf{17.9 } & \textbf{41.3 } & 52.9 \bigstrut[t]\\
    \hline
    action & \multicolumn{4}{c|}{waiting} & \multicolumn{4}{c|}{walkingdog} & \multicolumn{4}{c|}{walkingtogether} & \multicolumn{4}{c}{average} \bigstrut[t]\\
    \hline
    millisecond & 80   & 160  & 320  & 400  & 80   & 160  & 320  & 400  & 80   & 160  & 320  & 400  & 80   & 160  & 320  & 400 \bigstrut[t]\\
    \hline
    DMGNN & 12.2  & 24.2  & 59.6  & 77.5  & 47.1  & 93.3  & 160.1  & 171.2  & 14.3  & 26.7  & 50.1  & 63.2  & 17.0  & 33.6  & 65.9  & 79.7  \bigstrut[t]\\
    LTD  & 11.4  & 24.0  & 50.1  & 61.5  & 23.4  & 46.2  & 83.5  & 96.0  & 10.5  & 21.0  & 38.5  & 45.2  & 12.7  & 26.1  & 52.3  & 63.5  \bigstrut[t]\\
    SPGSN & 9.2  & 19.8  & 43.1  & 54.1  & —    & —    & —    & —    & 8.9  & 18.2  & 33.8  & 40.9  & 10.4  & 22.3  & 47.1  & 58.3  \bigstrut[t]\\
    PGBIG & 8.9  & 20.1  & 43.6  & 54.3  & 18.8  & 39.3  & 73.7  & 86.4  & 8.7  & 18.6  & 34.4 & 41.0 & 10.3  & 22.7  & 47.4  & 58.5  \bigstrut[t]\\
    \hline
    Ours & \textbf{7.6 } & \textbf{17.9 } & \textbf{41.1 } & \textbf{52.3 } & \textbf{16.0 } & \textbf{36.0 } & \textbf{72.0 } & \textbf{85.2 } & \textbf{7.7 } & \textbf{16.8 } & \textbf{32.3}  & \textbf{39.4}  & \textbf{9.1 } & \textbf{20.9 } & \textbf{45.9 } & \textbf{57.4}  \bigstrut[t]\\
    \hline
    \end{tabular}%
    \caption{Comparisons of short-term prediction for all actions and the average on H3.6M. The best results are shown in bold. Our method outperforms all baselines in most short-term prediction cases. }
  \label{h36_short}%
\end{table*}%

%% file: Table_h36_long.tex
\begin{table*}[t]
\small
\addtolength{\tabcolsep}{-3pt}
  \centering
    \begin{tabular}{c|cc|cc|cc|cc|cc|cc|cc|cc}
    \hline
    action & \multicolumn{2}{c|}{walking} & \multicolumn{2}{c|}{eating} & \multicolumn{2}{c|}{smoking} & \multicolumn{2}{c|}{discussion} & \multicolumn{2}{c|}{directions} & \multicolumn{2}{c|}{greeting} & \multicolumn{2}{c|}{phoning} & \multicolumn{2}{c}{posing} \bigstrut[t]\\
    \hline
    millisecond & 560  & 1000 & 560  & 1000 & 560  & 1000 & 560  & 1000 & 560  & 1000 & 560  & 1000 & 560  & 1000 & 560  & 1000 \bigstrut[t]\\
    \hline
    DMGNN & 73.4 & 95.8 & 58.1 & 86.7 & 50.9 & 72.2 & 81.9 & 138.3 & 110.1 & 115.8 & 152.5 & 157.7 & 78.9 & \textbf{98.6} & 163.9 & 310.1 \bigstrut[t]\\
    LTD  & 54.1 & 59.8 & 53.4 & 77.8 & 50.7 & 72.6 & 91.6 & 121.5 & 71   & 101.8 & 115.4 & 148.8 & 69.2 & 103.1 & 114.5 & 173 \bigstrut[t]\\
    SPGSN & \textbf{46.9} & \textbf{53.6} & \textbf{49.8} & \textbf{73.4} & 46.7 & 68.6 & —    & —    & 70.1  & 100.5  & —    & —    & 66.7  & 102.5  & —    & — \bigstrut[t]\\
    PGBIG & 48.1 & 56.4 & 51.1 & 76   & 46.5 & 69.5 & \textbf{87.1} & 118.2 & \textbf{69.3} & \textbf{100.4} & 110.2 & 143.5 & 65.9 & 102.7 & \textbf{106.1} & \textbf{164.8} \bigstrut[t]\\
    \hline
    Ours & 49.0  & 56.3  & 51.1  & 75.1  & \textbf{46.1 } & \textbf{68.2 } & \textbf{87.1 } & \textbf{117.2 } & 70.6  & 101.8  & \textbf{110.0 } & \textbf{141.7 } & \textbf{65.7 } & 101.9 & 109.4  & 165.8  \bigstrut[t]\\
    \hline
    action & \multicolumn{2}{c|}{purchases} & \multicolumn{2}{c|}{sitting} & \multicolumn{2}{c|}{sittingdown} & \multicolumn{2}{c|}{takingphoto} & \multicolumn{2}{c|}{waiting} & \multicolumn{2}{c|}{walkingdog} & \multicolumn{2}{c|}{walkingtogether} & \multicolumn{2}{c}{avearge} \bigstrut[t]\\
    \hline
    millisecond & 560  & 1000 & 560  & 1000 & 560  & 1000 & 560  & 1000 & 560  & 1000 & 560  & 1000 & 560  & 1000 & 560  & 1000 \bigstrut[t]\\
    \hline
    DMGNN & 118.6 & 153.8 & \textbf{60.1} & \textbf{104.9} & 122.1 & 168.8 & 91.6 & 120.7 & 106  & 136.7 & 194  & 182.3 & 83.4 & 115.9 & 103  & 137.2 \bigstrut[t]\\
    LTD  & 102  & 143.5 & 78.3 & 119.7 & 100  & 150.2 & 77.4 & 119.8 & 79.4 & 108.1 & 111.9 & 148.9 & 55   & 65.6 & 81.6 & 114.3 \bigstrut[t]\\
    SPGSN & —    & —    & 75.0  & 116.2  & —    & —    & 75.6  & \textbf{118.2 } & 73.5  & 103.6  & —    & —    & —    & —    & 77.4  & \textbf{109.6 } \bigstrut[t]\\
    PGBIG & \textbf{95.3} & \textbf{133.3} & 74.4 & 116.1 & \textbf{96.7} & \textbf{147.8} & \textbf{74.3} & 118.6 & \textbf{72.2} & \textbf{103.4} & 104.7 & 139.8 & 51.9 & 64.3 & \textbf{76.9} & 110.3 \bigstrut[t]\\
    \hline
    Ours & 96.9  & 137.5  & 74.7  & 116.6  & 96.9  & 150.2  & 77.3  & 120.8  & 73.3  & 104.1  & \textbf{103.8 } & \textbf{137.3 } & \textbf{51.5 } & \textbf{61.7 } & 77.5 & 110.4  \bigstrut[t]\\
    \hline
    \end{tabular}%
    \caption{Comparisons of long-term prediction for all actions and the average on H3.6M. Our method achieves competitive results with the best models. However, our model only has approximately 10$\%$ parameters as the best model.}
  \label{h36_long}%
\end{table*}%

%% file: Table_h36_stsgcn.tex
\begin{table}[t]
\small
  \centering 
    \begin{tabular}{c|cccccc}
    \hline
    millisecond & 80   & 160  & 320  & 400  & 560  & 1000 \bigstrut[t]\\
    \hline
    STSGCN & 10.1  & 17.1  & 33.1  & 38.3  & 50.8  & 75.6  \bigstrut[t]\\
    GAGCN & 10.1  & 16.9  & 32.5  & 38.5  & 50.0  & 72.9  \bigstrut[t]\\
    \hline
    Ours &   \textbf{6.3}  &  \textbf{11.9}    &   \textbf{24.1}   &  \textbf{30.2}    &  \textbf{42.4}    & \textbf{66.1} \bigstrut[t]\\
    \hline
    \end{tabular}%
    \caption{Comparison of prediction for the average on H3.6M under the evaluation criteria of STSGCN.}
   
  \label{stsgcn}%
\end{table}%

%% file: Table_cmu.tex
\begin{table}[htbp]
\small
  \centering
    \begin{tabular}{c|cccccc}
    \hline
    millisecond & 80   & 160  & 320  & 400  & 560  & 1000 \bigstrut[t]\\
    \hline
    DMGNN & 13.6  & 24.1  & 47.0  & 58.8  & 77.4  & 112.6  \bigstrut[t]\\
    LTD  & 9.3  & 17.1  & 33.0  & 40.9  & 55.8  & 86.2  \bigstrut[t]\\
    SPGSN & 8.3  & 14.8  & 28.6  & 37.0  &  $-$   & \textbf{77.8 } \bigstrut[t]\\
    PGBIG & 7.6  & 14.3  & 29.0  & 36.6 & 50.9 & 80.1  \bigstrut[t]\\
    \hline
    Ours & \textbf{7.2 } & \textbf{13.5 } & \textbf{27.9}  & \textbf{36.4}  & \textbf{48.6}  & 82.1  \bigstrut[t]\\
    \hline
    \end{tabular}%
    \caption{Comparison of prediction for the average on CMU}
  \label{cmu}%
\end{table}%

%% file: Table_3dpw.tex
\begin{table}[htbp]
\small
  \centering
    \begin{tabular}{c|cccccc}
    \hline
    millisecond & 100  & 200  & 400   & 600  & 800  & 1000 \bigstrut[t]\\
    \hline
    DMGNN & 17.8 & 37.1 & 70.4  & 94.1 & 109.7  & 123.9 \bigstrut[t]\\
    LTD  & 16.3 & 35.6 & 67.5  & 90.4 & 106.8 & 117.8 \bigstrut[t]\\
    SPGSN & 15.4 & 32.9 & 64.5  & 91.6 & 104.0  & 111.1 \bigstrut[t]\\
    $\text{PGBIG}^*$ & 13.1  & \textbf{29.2 } & 61.0    & 89.6  & 102.6   & 109.4 \bigstrut[t]\\
    \hline
    Ours & \textbf{12.4 } & \textbf{29.2 } & \textbf{59.1 }  & \textbf{87.7 } & \textbf{99.9 }& \textbf{107.7} \bigstrut[t]\\
    \hline
    \multicolumn{7}{l}{\scriptsize * This is a correction value that reproduced with original checkpoints} \\
    \end{tabular}%
    \caption{Comparison of prediction for the average on 3DPW}
  \label{3dpw}%
\end{table}%

%% file: Table_complexity.tex
\begin{table}[htbp]
  \centering
  \small
    \begin{tabular}{c|c|c|c|c}
    \hline
    Model & LTD   & SPGSN  & PGBIG  & Ours \bigstrut\\
    \hline
    Parameters & 2.55M & 5.67M & 1.74M & 0.55M \bigstrut\\
    \hline
    FLOPs & 133.7M & 549.3M & 55.8M & 143.5M \bigstrut\\
    \hline
    Inference Time & 5.18ms & 63.07ms & 16.22ms & 8.09ms  \bigstrut\\
    \hline
    \end{tabular}%
    \caption{Comparison of Computational Complexity}
  \label{complexity}%
\end{table}%

%% file: Tabel_jitter.tex
\begin{table}[t]
\small
  \centering
    \begin{tabular}{c|c|c|c}
    \hline
         & 0-1000ms & 400-1000ms & 800-1000ms \bigstrut\\
    \hline
    STSGCN & 671.76 & 518.99 & 649.05 \bigstrut\\
    \hline
    PGBIG & 161.20 & 195.91 & 227.81 \bigstrut\\
    \hline
    Ours & 111.82 & 114.70 & 119.24 \bigstrut\\
    \hline
    \end{tabular}%
    \caption{Comparison of Quality of Predicted Motion in Jitter Metric $(m/s^3)$}
  \label{jitter}%
\end{table}%

%% file: Table_ablation_model.tex
\begin{table}[htbp]
\small
  \centering
    \begin{tabular}{c|cccccc}
    \hline
         & 80   & 160  & 320  & 400  & 560  & 1000 \bigstrut[t]\\
    \hline
    GCN  &    9.7  &   22.2   &  47.9    &  59.7    & 78.4     & 111.8 \bigstrut[t]\\
    \hline
    SAGGB-3 & 9.3  & 22.0  & 48.2  & 60.1  & 79.5  & 112.6  \bigstrut[t]\\
    SAGGB-6 & \textbf{9.1 } & \textbf{20.9 } & \textbf{45.9 } & \textbf{57.4 } & \textbf{77.5 } & \textbf{110.4 } \bigstrut[t]\\
    SAGGB-8 & 9.1  & 20.9  & 46.4  & 58.3  & 77.7  & 110.9  \bigstrut[t]\\
    \hline
    \end{tabular}%
    \caption{Ablation on SAGGB module for network architecture, evaluated on long-term prediction on H3.6M}
  \label{ablation_model}%
\end{table}%